# Feature Level Fusion of Face and Palmprint Biometrics by Isomorphic Graph-based Improved K-Medoids Partitioning

Dakshina Ranjan Kisku[1], Phalguni Gupta[2] and Jamuna Kanta Sing[3]

[1] Department of Computer Science and Engineering,
Dr. B. C. Roy Engineering College,
Durgapur – 713206, India
[2] Department of Computer Science and Engineering,
Indian Institute of Technology Kanpur,
Kanpur – 208016, India
[3] Department of Computer Science and engineering,
Jadavpur University,
Kolkata – 700032, India
{drkisku, jksing}@ieee.org; pg@cse.iitk.ac.in

**Abstract.** This paper presents a feature level fusion approach which uses the improved K-medoids clustering algorithm and isomorphic graph for face and palmprint biometrics. Partitioning around medoids (PAM) algorithm is used to partition the set of n invariant feature points of the face and palmprint images into k clusters. By partitioning the face and palmprint images with scale invariant features SIFT points, a number of clusters is formed on both the images. Then on each cluster, an isomorphic graph is drawn. In the next step, the most probable pair of graphs is searched using iterative relaxation algorithm from all possible isomorphic graphs for a pair of corresponding face and palmprint images. Finally, graphs are fused by pairing the isomorphic graphs into augmented groups in terms of addition of invariant SIFT points and in terms of combining pair of keypoint descriptors by concatenation rule. Experimental results obtained from the extensive evaluation show that the proposed feature level fusion with the improved K-medoids partitioning algorithm increases the performance of the system with utmost level of accuracy.

## 1 Introduction

In multibiometrics fusion [1], feature level fusion [2,3] makes use of integrated feature sets obtained from multiple biometric traits. Fusion at feature level [2,3] is found to be useful than other levels of fusion such as match score fusion [4], decision fusion [4], rank level fusion [4]. Since feature set contains relevant and richer information about the captured biometric evidence, fusion at feature level is expected to provide more accurate authentication results. It is very hard to fuse multiple biometric evidences [2,3] at feature extraction level in practice because the feature sets are sometimes found to be incompatible. Apart from this reason, there are two more reasons to achieve fusion at feature extraction level such as the feature spaces are unknown for different biometric evidences and fusion of feature spaces may lead to the problem of curse of dimensionality problem [2]. Further, poor feature representation may cause to degrade the performance of recognition of users.

---
[1] Corresponding author

Multimodal systems [4] acquire information from more than one source. Unibiometric identifiers [5] use single source biometric evidence and often are affected by problems like lack of invariant representation, non-universality, noisy sensor data and lack of individuality of the biometric trait and susceptibility to circumvention. These problems can be minimized by using multibiometric systems that consolidate evidences obtained from multiple biometric sources. Feature level fusion [2] of biometric traits is a challenging problem in multimodal fusion. However, good feature representation and efficient solution to curse of dimensionality problem can lead to feature level fusion with ease.

Multibiometrics fusion [4] at match score level, decision level and rank level have extensively been studied and there exist a few feature level fusion approaches. However, to the best of the knowledge of authors, there is enough scope to design an efficient feature level fusion approach. The feature level fusion of face and palmprint biometrics proposed in [6] uses single sample of each trait. Discriminant features using graph-based approach and principal component analysis techniques are used to extract features from face and palmprint. Further, a distance separability weighting strategy is used to fuse two sets at feature extraction level. Another example of feature level fusion of face and hand biometrics has been proposed in [7]. It has been found that the performance of feature level fusion outperforms the match score fusion. In [8], a feature level fusion has been studied where phase congruency features are extracted from face and Gabor transformation is used to extract features from palmprint. These two feature spaces are then fused using user specific weighting scheme. A novel feature level fusion of face and palmprint biometrics has been presented in [9]. It makes use of correlation filter bank with class-dependence feature analysis method for feature fusion of these two modalities.

A feature level fusion of face [10] and palmprint [11] biometrics using isomorphic graph [12] and K-medoids [13] is proposed in this paper. SIFT feature points [14] are extracted from face and palmprint images as part of feature extraction work. Using the partitioning around medoids (PAM) algorithm [15] which is considered as a realization of K-medoids clustering algorithm is used to partition the face and palmprint images of a set of $n$ invariant feature points into $k$ number of clusters. Then for each cluster, an isomorphic graph is drawn on SIFT points which belong to the clusters. Graphs are drawn on each partition or cluster by searching the most probable isomorphic graphs using iterative relaxation algorithm [16] from all possible isomorphic graphs while the graphs are compared between face and palmprint templates. Each pair of clustered graphs are then fused by concatenating the invariant SIFT points and all pairs of isomorphic graphs of clustered regions are further fused to make a single concatenated feature vector. The same set of invariant feature vector is also constructed from query pair of samples of face and palmprint images. Finally, matching between these two feature vectors is determined by computing the distance using K-Nearest Neighbor [17] and normalized correlation [18] distance approaches. IIT Kanpur multimodal database is used for evaluation of the proposed feature level fusion technique.

The paper is organized as follows. Next section discusses SIFT features extraction from face and palmprint images. Section 3 presents K-Medoids partitioning of SIFT features into a number of clusters. The method of obtaining isomorphic graphs on the sets of the SIFT points which belong to the clusters is also discussed in the same section. Next section presents feature level fusion of clustered SIFT points by pairing two graphs of a pair of clustered regions drawn on face and palmprint images. Experimental results are presented in Section 5 while conclusion is made in the last section.

## 2   SIFT Keypoints Extraction

To recognize and classify objects efficiently, feature points from objects can be extracted to make a robust feature descriptor or representation of the objects. David Lowe [14] has introduced a technique to extract features from images which are called Scale Invariant Feature Transform (SIFT). These features are invariant to scale, rotation, partial illumination and 3D projective transform and they are shown to provide robust matching across a substantial range of affine distortion, change in 3D viewpoint, addition of noise, and change in illumination. SIFT image features provide a set of features of an object that are not affected by occlusion, clutter and unwanted noise in the image. In addition, the SIFT features are highly distinctive in nature which have accomplished correct matching on several pair of feature points with high probability between a large database and a test sample. Following are the four major filtering stages of computation used to generate the set of features based on SIFT [14].

In the proposed work, the face and palmprint images are normalized by adaptive histogram equalization [2]. Localization of face is done by the face detection algorithm proposed in [19] while localization of palmprint is made by the algorithm discussed in [20]. After geometric normalization and spatial enhancement, SIFT features [14] are extracted from the face and palmprint images. Each feature point is composed of four types of information – spatial location ($x$, $y$), scale ($S$), orientation ($\theta$) and Keypoint descriptor ($K$). For the experiment, only keypoint descriptor [14] information has been considered which consists of a vector of 128 elements representing neighborhood intensity changes of each keypoint. More formally, local image gradients are measured at the selected scale in the region around each keypoint. The measured gradients information is then transformed into a vector representation that contains a vector of 128 elements for each keypoint calculated over extracted keypoints. These keypoint descriptor vectors represent local shape distortions and illumination changes. In Figure 1 and Figure 2, SIFT features extractions are shown for the face and palmprint images respectively.

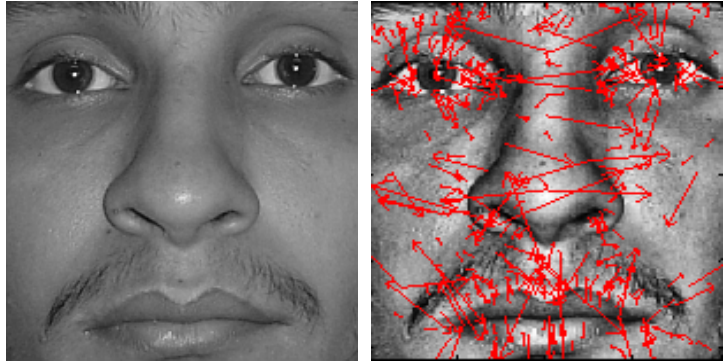

**Fig. 1.** Face Image and SIFT Keypoints Extraction

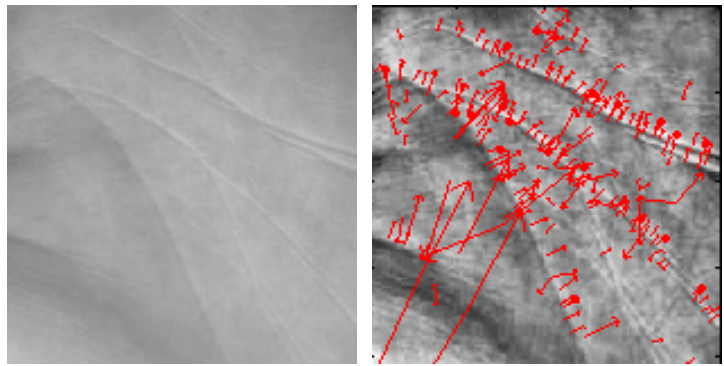

**Fig. 2.** Palm Image and SIFT Keypoints Extraction

## 3   Feature Partitioning and Isomorphic Graph Representation

In most multimodal biometric applications [4], lack of well feature representation leads to the degradation of the performance. Therefore, well representation of feature space and template in terms of invariant feature points may help to exhibit robust and efficient solution towards user authentication. In spite of considering the whole biometric template and all the SIFT keypoints, clustering of all feature points into a number of clusters with limited number of invariant points can be an efficient approach of feature space representation. Clustering approach [21] often gathers together the keypoints which are most relevant and useful members of a particular cluster and association of these keypoints represents the relation within the keypoints in a cluster.

The proposed fusion approach partition the SIFT keypoints [14] which are extracted from face and palmprint images into a number of clustering regions with limited number of keypoints in each cluster and then isomorphic graph [12] is formed on each cluster with the keypoints of partitioned face and palmprint images. Prior to construct the isomorphic graphs on clusters, corresponding pairs of clusters are established in terms of relation between keypoints and geometric distance between keypoints regarded as vertices and edges respectively to itself as auto-isomorphism [12] for face and palmprint images. Three different steps are followed to make a correspondence between a pair of face cluster and a palmprint cluster after clustering of keypoints. Since the number of keypoints on face is more than that on palmprint, face image can be made as reference with respect to palmprint image. Later auto-isomorphism graph is built on the each face cluster with the keypoints and the corresponding isomorphism is built on a palm cluster while point correspondences are established using point pattern matching approach [3]. Then a pair of clusters corresponding to a pair of face and palmprint images is searched by mapping the isomorphic graph of face cluster to the isomorphic graph of palmprint cluster. This process is carried out for all pairs of clusters of face and palmprint images. Lastly, the fusion of each pair of clusters of identical dimension of keypoints is dome by sum rule approach [3]. Since each keypoint descriptor is a vector of 128 elements and each face and palm cluster is represented by an isomorphic graph. Isomorphic graphs for both the face and palm clusters contain same number of keypoints with one-to-one mapping. These two feature vectors containing SIFT keypoints are then fused using sum rule.

## 3.1 SIFT Keypoints Partitioning using PAM Characterized K-Medoids Algorithm

A medoid can be defined as the object of a cluster, which means dissimilarity to all the objects in the cluster is minimal. K-medoids [13] chooses data points as cluster centers (also called 'medoids'). K-medoids clusters the dataset of *n* objects into *k* clusters and is more robust to noise and outliers as compared to K-means clustering algorithm. This clustering algorithm is an adaptive version of K-means clustering approach and is used to partition the dataset into a number of groups which minimizes the squared error between the points that belong to a cluster and a point designated as the center of the cluster. The generalization of K-medoids algorithm is the Partitioning around Medoids (PAM) algorithm [15] which is applied to the SIFT keypoints of face and palmprint images to obtain the partitioned of features which can provide more discriminative and meaningful clusters of invariant features. The algorithm can be given below.

*Step 1: Select randomly k number of points from the SIFT points set as the medoids.*
*Step 2: Assign each SIFT feature point to the closest medoid which can be defined by a distance metric (i.e., Minkowski distance over the Euclidean space)*
*Step 3: for each medoid i, i = 1, 2...k*

*for each non-medoid SIFT point j*
  *swap i and j and*
  *compute the total cost of the configuration*
Step 4: Select the configuration with the lowest cost
Step 5: Repeat Step 2 to Step 5 until there is no change in the medoid.

**Improved version of PAM clustering using Silhouette approximations.** Silhouette technique [15] can be used to verify the quality of a cluster of data points. After applying the PAM clustering technique [15] to the sets of SIFT keypoints for face and palmprint images, each cluster can be verified by Silhouette technique. Let, for each keypoint $i$, $x(i)$ be the average distance of $i$ with all the keypoints in cluster $c_m$. Consider $x(i+1)$ as an additional average distance next to $x(i)$. These two successive distances $x(i)$ and $x(i+1)$ are considered to verify the matching of these keypoints $i$ and $(i+1)$ to the cluster where these points are assigned. Then the average distances of $i$ and $(i+1)$ with the keypoints of another single cluster are found. Repeat this process for every cluster in which $i$ and $(i+1)$ are not a member. If the cluster with lowest average distances to $i$ and $(i+1)$ are $y(i)$ and $y(i+1)$ ($y(i+1)$ is the next lowest average distance to $y(i)$), the cluster is known to be the neighboring cluster of the former cluster in which $i$ and $(i+1)$ are assigned. It can be defined by the following equation

$$S(i) = \frac{(y(i) + y(i+1))/2 - (x(i) + x(i+1))/2}{\max[((x(i) + x(i+1)), (y(i) + y(i+1))]} \tag{1}$$

From Equation (1) it can be written that $-1 \leq S(i) \leq 1$

When $x(i)+x(i+1) << y(i)+y(i+1)$, $S(i)$ would be very closer to 1. Distances $x(i)$ and $x(i+1)$ are the measures of dissimilarity of $i$ and $(i+1)$ to its own cluster. If $y(i)+y(i+1)$ is small enough, then it is well matched, otherwise when the value of $y(i)+y(i+1)$ is large then bad match is occurred. Keypoint is well clustered when $S(i)$ is closer to 1 and when that value of $S(i)$ is negative then it belongs to another cluster. $S(i)$ zero means keypoint is on the border of any two clusters.

The existing algorithm has been extended by taking another average distances $x(i+1)$ and $y(i+1)$ for a pair of clusters and it has been determined that a better approximation could be arise while PAM algorithm is used for partition the keypoints set. The precision level of each cluster is increased by this improved approximation method where more relevant keypoints are taken instead of taking restricted number of keypoints for fusion.

### 3.2 Establishing Correspondence between Clusters of Face and Palmprint Images

To establish correspondence [10] between any two clusters of face and palmprint images, it has been observed that more than one keypoint on face image may correspond to single keypoint on the palmprint image. To eliminate false matches and to consider the only minimum pair distance from a set of pair distances for making correspondences, first it needs to verify the number of feature points that are available in the cluster of face and that in the cluster of palmprint. When the number of feature points in the cluster for face is less than that of the cluster for palmprint, many points of interest from the palmprint cluster needs to be discarded. If the number of points of interest on the face cluster is more than that of the palmprint cluster, then a single interest point on the palmprint cluster may act as a match point for many points of interest of face cluster. Moreover, many points of interest on the face cluster may have correspondences to a single point of interest on the cluster for palmprint. After computing all distances between points of interest of face cluster and palmprint cluster that have made correspondences, only the minimum pair distance is paired.

After establishing correspondence between clusters for face and palmprint images, isomorphic graph representation [12] for each cluster has been formed while removing few more keypoints from the paired clusters. Further iterative relaxation algorithm [16] is used for searching the best possible pair of isomorphic graphs from all possible graphs.

### 3.3 Isomorphic Graph Representations of Partitioned Clusters

To interpret each pair of clusters for face and palmprint, isomorphic graph representation has been used. Each cluster contains a set of SIFT keypoints [14] and each keypoint is considered as a vertex of the proposed isomorphic graph. A one-to-one mapping function is used to map the keypoints of the isomorphic graph constructed on a face cluster to a palmprint cluster while these two clusters have been made correspondence to each other. When two isomorphic graphs are constructed on a pair of face and palmprint clusters with equal number of keypoints, two feature vectors of keypoints are constructed for fusion.

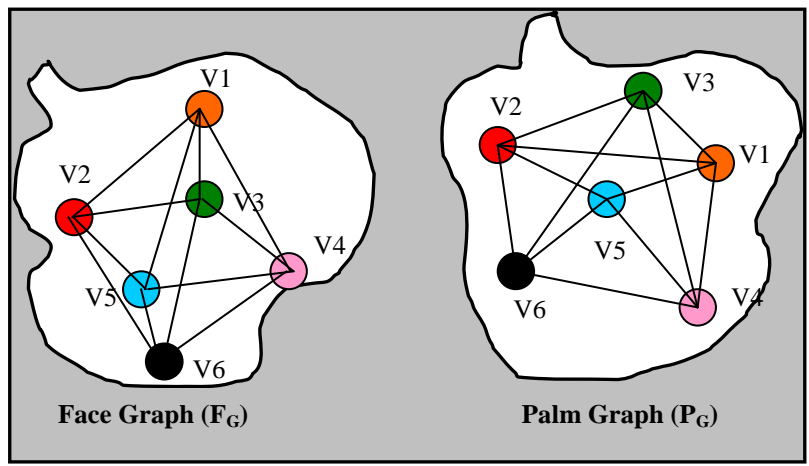

**Fig. 3.** One-to-one Correspondence between Two Isomorphic Graphs

Let $F_G$ and $P_G$ be two graphs and also let $f$ be a mapping function from the vertex set of $F_G$ to vertex set of $P_G$. So when

- $f$ is one-to-one and
- $f(v_k)$ is adjacent to $f(w_k)$ in $P_G$ if and only if $v_k$ is adjacent to $w_k$ in $F_G$

Then the function $f$ is known as an isomorphism and two graphs $F_G$ and $P_G$ are isomorphic. Therefore the two graphs $F_G$ and $P_G$ are isomorphic if there is a one-to-one correspondence between vertices of $F_G$ and those of $P_G$ while two vertices of $F_G$ are adjacent then so are their images in $P_G$. If two graphs are isomorphic then they are identical graph though the location of the vertices may be different. Figure 3 shows an example of isomorphic graph and one-to-one correspondence between two isomorphic graphs where each colored circle refers independent vertex.

## 4. Fusion of keypoints and matching

### 4.1 Fusion of Keypoints

To fuse the SIFT keypoint descriptors obtained from each isomorphic graph for face and for palmprint images, two different fusion rules are applied serially, viz. sum rule [3] and concatenation rule [2]. Let $F_G(v_k) = (v_{k1}, v_{k2}, v_{k3},..., v_{kn})$ and $P_G(w_k) = (w_{k1}, w_{k2}, w_{k3}, ..., w_{kn})$ be the two sets of keypoints obtained from two isomorphic graphs for a pair of face and palmprint clusters. Suppose there are $m$ numbers of clusters in each of face and palmprint images. Then these two sets of clusters can be fused using sum fusion rule and the concatenation rule can be further applied to form an integrated feature vector. Suppose that $F_{G1}$, $F_{G2}$, $F_{G3}$, ..., $F_{Gm}$ sets of keypoints are obtained from a face image after clustering and isomorphism and $P_{G1}$, $P_{G2}$, $P_{G3}$, ..., $P_{Gm}$ are the sets of keypoints obtained from a palmprint image. The sum rule can be defined for the fusion of keypoints as follows

$$S_{FP1} = F_{G1} + P_{G1} = \{(v_{k1}^1 + w_{k1}^1), (v_{k2}^1 + w_{k2}^1), (v_{k3}^1 + w_{k3}^1),...,(v_{kn}^1 + w_{kn}^1)\}$$

$$S_{FP2} = F_{G2} + P_{G2} = \{(v_{k1}^2 + w_{k1}^2), (v_{k2}^2 + w_{k2}^2), (v_{k3}^2 + w_{k3}^2),...,(v_{kn}^2 + w_{kn}^2)\} \quad (2)$$

----- ----- ----- ----- ----- ----- ----- ----- ----- ----- -----
----- ----- ----- ----- ----- ----- ----- ----- ----- ----- -----
----- ----- ----- ----- ----- ----- ----- ----- ----- ----- -----

$$S_{FPm} = F_{Gm} + P_{Gm} = \{(v_{k1}^m + w_{k1}^m), (v_{k2}^m + w_{k2}^m), (v_{k3}^m + w_{k3}^m),...,(v_{kn}^m + w_{kn}^m)\}$$

In Equation (2), $S_{FPj}$ ($i = 1,2,...,m$), $v_{kj}$ ($j = 1,2,...,n$) and $w_{kj}$ ($j = 1,2,...,n$) refer to a fused set of keypoint descriptors for a pair of isomorphic graphs obtained by applying sum fusion rule, a keypoint of a face graph and a keypoint of a palm graph respectively. In the next step, concatenation rule is applied to the sets of keypoints to form a single feature vector.

### 4.2 Matching Criterion and Verification

The K-Nearest Neighbor (K-NN) distance [17] and correlation distance [18] approaches are used to compute distances from the concatenated feature sets. In K-NN approach, Euclidean distance metric is used to get K best matches. Let $d_i$ be the Euclidian distance of the concatenated feature set of subject $S_i$, $i = 1, 2, .... K$, which belong to the $K$ best matches against a query subject. Then the subject $S_t$ is verified against the query subject if $d_t \leq Th$ where $d_t$ is the minimum of $d_1, d_2, ..., d_K$ and $Th$ is the pre-assigned threshold.

On the other hand, the correlation distance metric is used for computing distance between a pair of reference fused feature set and probe fused feature set. The similarity between two concatenated feature vectors $f_1$ and $f_2$ can be computed as follows

$$d = \frac{\sum f_1 f_2}{\sqrt{\sum f_1 \sum f_2}} \tag{3}$$

Equation (3) denotes the normalized correlation between feature vectors $f_1$ and $f_2$. Let $d_i$ be the similarity of the concatenated feature set of subject $S_i$, $i = 1, 2, … K$, with respect to that of a query subject. Then the subject $S_t$ is verified against the query subject if $d_t \geq Th$ where $d_t$ is the maximum of $d_1, d_2, ..., d_K$ and $Th$ is the pre-assigned threshold.

## 5. Experimental results and discussion

### 5.1 Database

The proposed feature level fusion approach is tested on IIT Kanpur multimodal database containing face and palmprint images of 400 subjects each. To conduct experiment two face and two palmprint images are taken for each subject.

Face images are taken in controlled environment with maximum tilt of head by 20 degree from the origin. However, for evaluation purpose frontal view faces are used with uniform lighting and minor change in facial expression. These face images are acquired in two different sessions. Among the two face images, one image is used as a reference face and the other one is used as a probe face. After preprocessing of face images, it is then cropped by taking the face portion only for evaluation. To detect the face portion efficiently, the algorithm for face detection discussed in [19] is used.

Palmprint images are also taken in controlled environment with a flat bed scanner having spatial resolution of 200 dpi. Palmprint impressions are taken on the scanner with rotation of at most $\pm 35^0$ to each user. 800 palmprint images are collected from 400 subjects and each subject is contributed 2 palmprint images. Palmprint images are preprocessed with an image enhancement technique to achieve uniform spatial resolution. In the next step, palm portion is detected from palmprint image and this is achieved by the technique proposed in [20].

### 5.2 Experimental Results

The performance of the proposed fusion approach is determined using one-to-one matching strategy. Experimental results are obtained using two distance approaches namely, K-Nearest Neighbor (K-NN) distance [17] and normalized correlation [18]. We have not only determined the performance of the proposed feature level fusion approach, but also that of face and palmprint modality independently. Fused feature set which is obtained from reference face and palmprint images is matched with the feature set obtained from probe pair of face and palmprint images by computing the distance between these two sets. The Receiver Operating Characteristic curves (ROC) are determined for the six distinct cases: (i) face modality using K-NN, (ii) face modality using normalized correlation method, (iii) palmprint modality using K-NN, (iv) palmprint modality using normalized correlation method, (v) feature fusion verification using K-NN and (vi) feature fusion verification using normalized correlation method.

**Table 1.** Different Error Rates

| METHOD | FAR (%) | RECOGNITION RATE (%) |
|---|---|---|
| Face Recognition (K-NN) | 7.0 | 92.50 |
| Face Recognition (Correlation) | 6.0 | 93.75 |
| Palmprint Verification (K-NN) | 4.5 | 94.75 |
| Palmprint Verification (Correlation) | 2.5 | 96.00 |
| Feature Level Fusion (K-NN) | 2.0 | 97.50 |
| Feature Level Fusion (Correlation) | 0.0 | 98.75 |

False Accept Rate (FAR), False Reject Rate (FRR) and recognition rate are determined from the 800 face and palmprint images of 400 subjects. Feature level fusion method using normalized correlation outperforms other proposed methods including individual matching of face and palmprint modalities. The correlation metric based feature level fusion obtained 98.75% recognition rate with 0% FAR while K-NN based feature level fusion method has the recognition rate of 97.5% with 2% FAR. It can also be noted that FAR all the proposed methods are found to be less than its corresponding FRR. On the other hand, palmprint modality performs better than face modality while K-NN and correlation metrics are used. The distance metrics play an important role irrespective of use of invariant features and isomorphic graphs representations. However, the robust representation of face and palmprint images using isomorphic graphs with use of invariant SIFT keypoints and PAM characterized K-Medoids algorithm makes use of the proposed feature level fusion method to be efficient one. In single modality, the same approach has been used which are applied in feature level fusion method. Therefore, the different error rates obtained from the single modalities and feature fusion method are determined under a uniform framework. However, the methodology used for feature level fusion found to be not only superior to other methods and also shows significant improvements in terms of recognition rate and FAR. Table 1 shows different error rates for the proposed methods while the Receiver Operating Characteristics (ROC) curves determined at different operating threshold points are given in Figure 4.

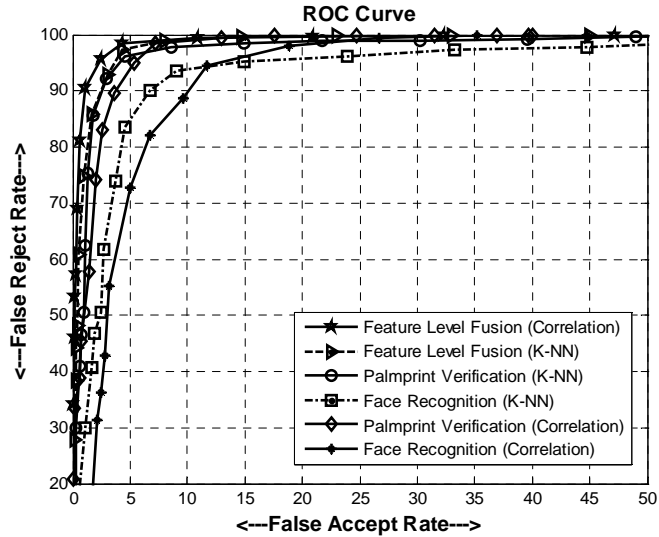

**Fig. 4.** Receiver Operating Characteristics (ROC) Curves

## 6. Conclusion

This paper has presented a feature level fusion system of face and palmprint traits using invariant SIFT descriptor and isomorphic graph representation. The performance of feature level fusion has verified by two distance metrics namely, K-NN and normalized correlation metrics. However, normalized correlation metric is found to be superior to that of K-NN metric for all the verification methods proposed in this paper. The main aim of this paper is to present a robust representation to invariant SIFT features for face and palmprint images which cannot only be useful to the individual verification of face and palmprint modality but has also proved to be useful to the proposed feature level fusion approach. Since isomorphic graph is used for representation of feature points extracted from face and palmprint images, therefore identical numbers of matched pair points are to be used for fusion. In addition PAM characterized K-Medoids algorithm as a feature reduction technique has also proved to be useful for keeping relevant nature of feature keypoints. Single modality palmprint method performs better than face modality while K-NN and correlation approaches are used. Feature level fusion approach attains 98.75% recognition rate at 0% FAR and can also be used efficiently.